\documentclass[10pt,twocolumn,letterpaper]{article}

\usepackage[pagenumbers]{cvpr} 

\definecolor{cvprblue}{rgb}{0.21,0.49,0.74}
\usepackage[accsupp]{axessibility}
\usepackage[pagebackref,breaklinks,colorlinks,allcolors=cvprblue]{hyperref}
\usepackage{multirow, multicol}
\usepackage{wrapfig}
\usepackage{graphicx}
\definecolor{oursbg}{HTML}{F5F2F7}
\definecolor{sbred}{HTML}{DB6057}
\definecolor{sbgreen}{HTML}{92DB58}
\definecolor{sbblue}{HTML}{5771DB}
\definecolor{sbpurple}{HTML}{A157DB}

\newcommand{\cc}{\cellcolor{oursbg}}

\def\confName{CVPR}
\def\confYear{2025}

\title{CityWalker: Learning Embodied Urban Navigation from Web-Scale Videos}

\author{Xinhao Liu\thanks{Equal contribution.} \quad Jintong Li\footnotemark[1]  \quad Yicheng Jiang \quad 
 Niranjan Sujay \quad Zhicheng Yang \quad \\ Juexiao Zhang \quad John Abanes \quad Jing Zhang \quad Chen Feng\thanks{Corresponding author}\\
New York University\\
{\small \bf \url{https://ai4ce.github.io/CityWalker/}}
}

\begin{document}
\twocolumn[{
\renewcommand\twocolumn[1][]{#1}
\maketitle
\vspace{-7mm}
\includegraphics[width=0.99\textwidth]{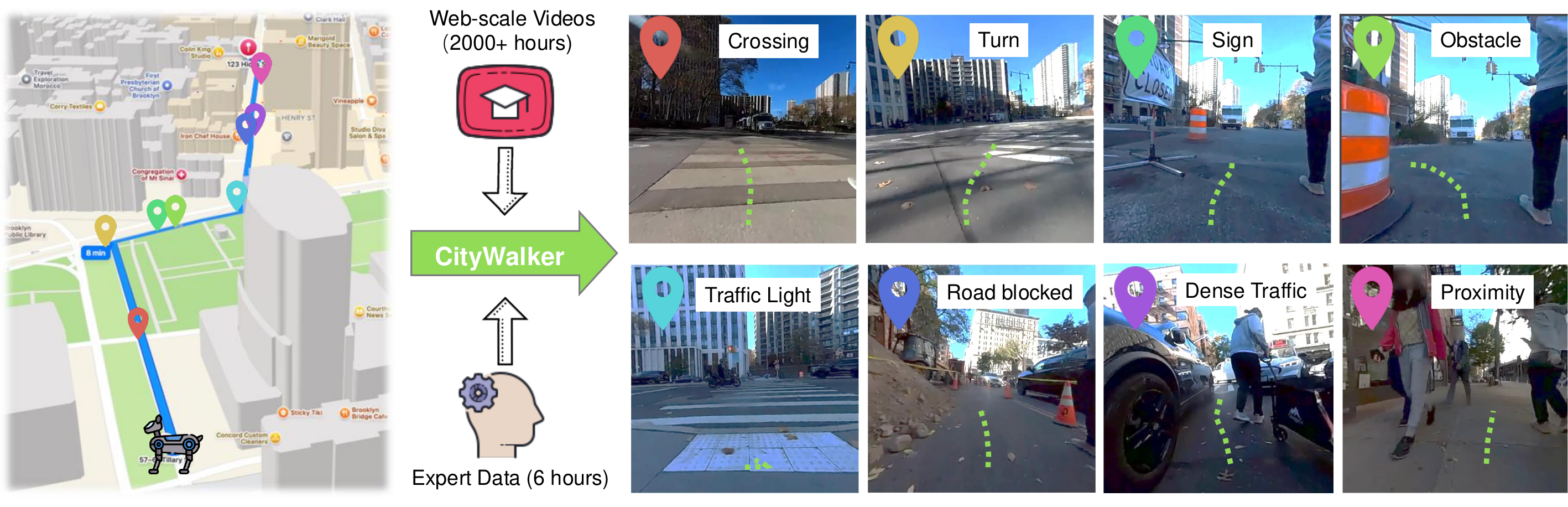}
\vspace{-2mm}
\captionof{figure}{\textbf{Embodied Urban Navigation}. Navigating urban spaces is challenging for (especially off-street) mobile agents. The differently colored pins~(\includegraphics[height=1em]{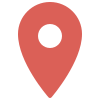}) along the route highlight various critical scenarios unique to complex and dynamic urban landscapes. Thumbnails on the right with corresponding colored pins demonstrate the real-world observation of these challenging cases. Our CityWalker model is trained with over 2000 hours of city walking videos and fine-tuned with a small amount of expert data to address these challenges effectively.}
\label{fig:teasing}
\vspace{2mm}
}]
{
  \renewcommand{\thefootnote}%
    {\fnsymbol{footnote}}
  \footnotetext[1]{Equal contribution ({\tt \{xinhao.liu, jintong.li\}@nyu.edu})}
  \footnotetext[2]{Corresponding author{\tt (cfeng@nyu.edu)}}
}
\vspace{-3mm}
\begin{abstract}
Navigating dynamic urban environments presents significant challenges for embodied agents, requiring advanced spatial reasoning and adherence to common-sense norms. Despite progress, existing visual navigation methods struggle in map-free or off-street settings, limiting the deployment of autonomous agents like last-mile delivery robots. To overcome these obstacles, we propose a scalable, data-driven approach for human-like urban navigation by training agents on thousands of hours of in-the-wild city walking and driving videos sourced from the web. We introduce a simple and scalable data processing pipeline that extracts action supervision from these videos, enabling large-scale imitation learning without costly annotations. Our model learns sophisticated navigation policies to handle diverse challenges and critical scenarios. Experimental results show that training on large-scale, diverse datasets significantly enhances navigation performance, surpassing current methods. This work shows the potential of using abundant online video data to develop robust navigation policies for embodied agents in dynamic urban settings.
\vspace{-7mm}
\end{abstract}    
\section{Introduction}
\label{sec:intro}
Visual navigation is a crucial capability for mobile agents. In previously unseen environments, humans typically rely on navigation tools (such as Google Maps) to reach a goal location by following a series of waypoints. These tools provide high-level guidance, but the actual navigation between waypoints requires sophisticated spatial awareness and decision-making. In urban settings, this involves understanding and adhering to complex rules, such as common sense regulations and norms, while dynamically responding to obstacles and environmental changes.
The diversity and complexity encountered in urban navigation scenarios pose significant challenges for navigation policies but are essential for deploying mobile robots in real-world settings. Applications like delivery robots and robotaxis, especially without high-definition maps, highlight the growing need for embodied agents to navigate efficiently and safely in dynamic urban landscapes. 
\textit{How can we teach embodied agents to navigate dynamic urban landscapes as effectively as humans do?}

The question has been explored in early works~\cite{muhlbauer2009navigation,kummerle2013navigation,morales2009autonomous} before the deep learning era. These approaches combined SLAM-based mapping with modular systems for searching, detection, analysis, and planning. The development of high-performance photorealistic simulators~\cite{savva2019habitat,xia2018gibson,kolve2017ai2,chang2017Matterport} significantly advanced research in visual navigation by enabling reinforcement learning (RL) in static indoor environments, achieving near-perfect point-goal navigation~\cite{Wijmans2019DDPPO}. This simulation-RL paradigm was also extended to more complex settings, including real-world panoramic street views~\cite{mirowski2018learning,li2019cross,hermann2020learning,mirowski2019streetlearn}. Recent works begin to explore imitation learning from expert demonstrations for long-horizon or cross-embodiment navigation tasks~\cite{shah2022viking,shah2023gnm}.

Despite their success in simulators or moderately complex environments, existing methods fall short in addressing the complexity of \textbf{embodied urban navigation}. Public urban spaces are inherently dynamic and unpredictable, \textit{characterized by ``multifarious terrains, diverse obstacles, and dense pedestrians'' that require real-time adaptability}~\cite{wu2024metaurban}. Moreover, effective navigation in such settings demands \textit{adherence to common sense navigation rules and social norms, such as using sidewalks, obeying traffic signals, and maintaining appropriate personal space to avoid conflicts}~\cite{sathyamoorthy2021comet,sun2021move,onozuka2021autonomous}. Current reinforcement learning and imitation learning approaches typically excel in static or controlled environments but struggle with real-world urban navigation. This is because these nuanced behaviors and delicate constraints are challenging to incorporate into simulation environments while ensuring reliable and generalizable sim-to-real transfer. Moreover, coverage of these complexities is still lacking in existing teleoperation demonstration datasets.
As a result, embodied urban navigation remains an \textbf{unsolved problem}.

While a straightforward approach to the problem is to collect expert trajectories through teleoperation, this is limited by data size and lack of diversity, which weakens the agent’s ability to generalize across various urban scenarios. 
\textit{Is there a data-driven approach that requires minimum annotation but maximizes generalizability?}
Inspired by the success of scaling laws in language, vision, and robotics tasks~\cite{brown2020language,kirillov2023segment,oquab2024dinov,driess2023palm}, we propose a scalable framework, \textbf{CityWalker}, which leverages web-scale city walking and driving videos to train models for embodied urban navigation. Our model is trained on over two thousand hours of internet-sourced videos across different geolocations, weather conditions, and times of day.

The subsequent question is \textit{how to effectively obtain action supervision from in-the-wild videos for imitation learning}. Current solutions rely heavily on prompting proprietary VLMs to generate \textit{action} labels from videos~\cite{wang2024vlm,hirose2024lelan}, which is costly and challenging to control, making it difficult to scale for large datasets. We argue that noisy pseudo labels from off-the-shelf visual odometry (VO) models~\cite{teed2024deep} are sufficiently effective for imitation learning in embodied urban navigation. Moreover, this data processing method generalizes beyond walking data. We demonstrate that a model trained with driving videos also performs well on quadruped agents, with further improvements observed when training in a cross-domain setting that combines both city walking and driving data.

We summarize our contribution as the following:

\begin{itemize}
    \item We identify embodied urban navigation as an intriguing yet unsolved problem. To address this problem, we propose a scalable, data-driven solution by leveraging web-scale city walking and driving videos.
    \item We introduce a simple yet scalable data processing paradigm,  making large-scale imitation learning feasible for urban navigation without extensive manual labeling. 
    \item We demonstrate training on web-scale data significantly improves navigation performance in real-world experiments, allowing embodied agents to handle the complexities of urban scenarios. We also show model performance scales well with training data size.
\end{itemize}

\section{Related Work}
\label{sec:related-work}

\textbf{Navigation in Simulation.} The introduction of high-performance photorealistic simulators~\cite{savva2019habitat,xia2018gibson,kolve2017ai2,chang2017Matterport} has significantly boosted navigation research in indoor household environments. Based on the form of goal specification, navigation tasks are usually categorized into point-goal~\cite{jaderberg2017reinforcement,gupta2017cognitive,savva2019habitat,Wijmans2019DDPPO}, image-goal~\cite{savinov2018semiparametric,hahn2021no,kwon2021visual,krantz2023navigating}, object-goal~\cite{gervet2023navigating,chaplot2020object,majumdar2022zson}, and vision-language navigation~\cite{das2018embodied,tan2019learning,krantz2020beyond}. While tasks like image-goal and object-goal navigation focus on efficient goal-finding or instruction understanding, point-goal navigation has a more straightforward goal specification and emphasizes safe and efficient trajectory planning to the goal. Our problem formulation resembles point-goal navigation in that accurate waypoints (represented as GPS coordinates) are readily available from navigation tools. Recently, point-goal navigation has been considered a "solved" problem due to near-perfect performance achieved in simulators~\cite{Wijmans2019DDPPO}. However, our investigation shows that waypoint-goal navigation in public urban spaces remains an unsolved challenge.

\begin{figure*}[t]
    \centering
    \includegraphics[width=0.9\linewidth]{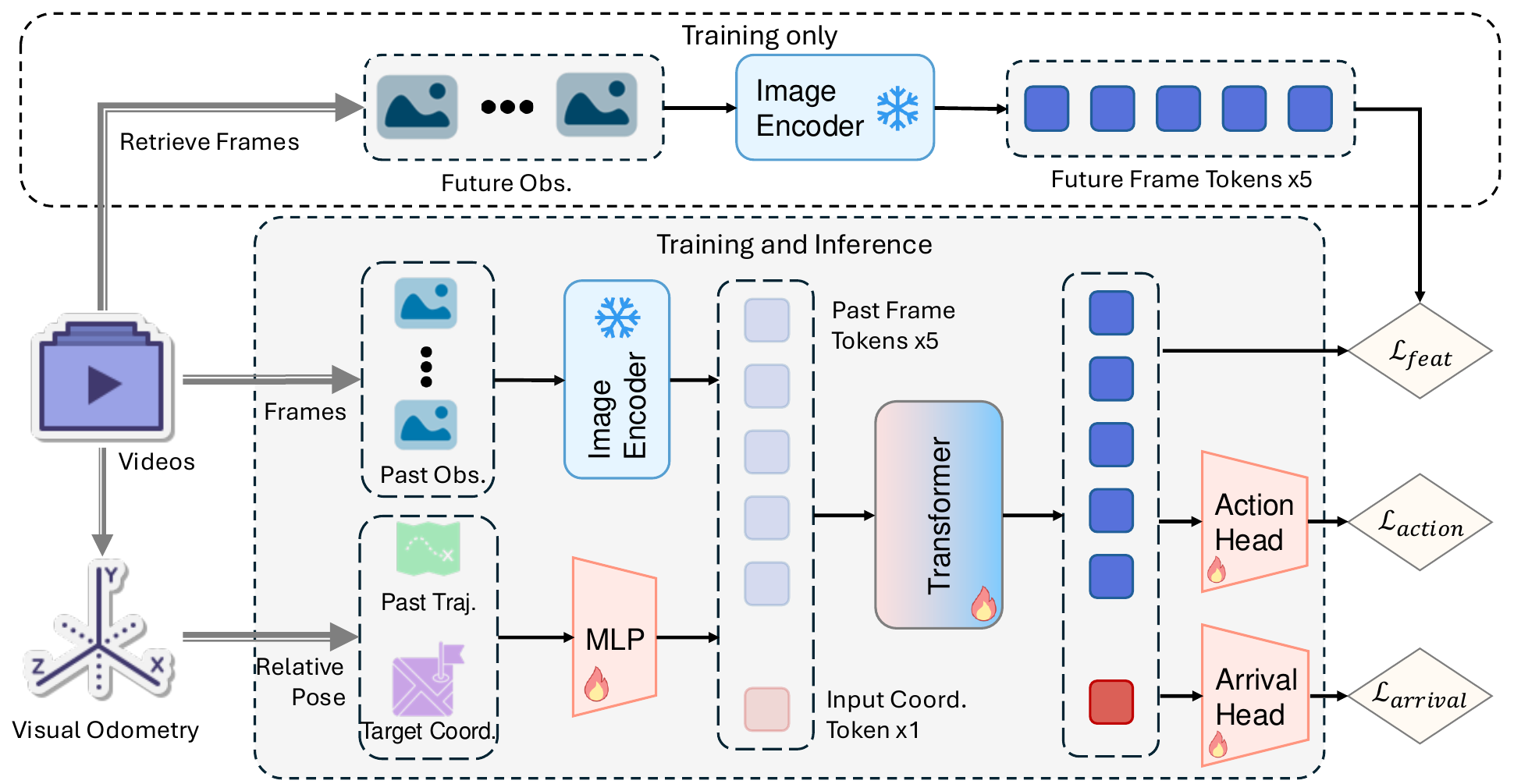}
    \caption{\textbf{Overall Illustration of CityWalker.} Our training pipeline starts with internet-sourced videos, using visual odometry to obtain relative poses between frames. At each time step, the model receives past observations, past trajectory, and target location as input. They are encoded via a frozen image encoder and a trainable coordinate encoder. A transformer processes these inputs to generate future tokens. An action head and an arrival head decode these tokens into action and arrival status predictions. During training, future frame tokens from future frames guide the transformer to hallucinate future tokens.}
    \label{fig:pipeline}
    \vspace{-4mm}
\end{figure*}

\noindent\textbf{Real-World Navigation.} Real-world robot navigation has a long history, predating the deep learning era. Early methods relied on ranging sensors and modular SLAM systems for navigation~\cite{muhlbauer2009navigation,kummerle2013navigation,morales2009autonomous}. Witnessing the success in simulators, various methods have been proposed to bridge the sim-to-real gap~\cite{sorokin2022learning,truong2023rethinking,truong2021learning,liu2024x}. Another thread of work learns navigation policies from expert trajectories~\cite{kahn2021badgr,kahn2021land}. By incorporating data from different embodiments~\cite{shah2023gnm} and carefully designed heuristics~\cite{shah2022viking}, these methods have been extended to long-horizon navigation on different robots~\cite{shah2023vint,sridhar2024nomad}. While deployed on real-world embodied agents, these methods still operate in relatively simple and static environments such as suburban areas, parking garages, or controlled obstacle courses. They are not guaranteed to generalize to significantly different, dynamic, and complex urban landscapes. Our work follows this line of research by training navigation models with imitation learning. But we focus on learning the rules and norms essential for urban environments, which is crucial for enabling applications such as delivery robots and robo-taxis.

\noindent\textbf{In-the-Wild Video Learning.} Learning from web-scale data has proven successful in language and vision tasks~\cite{brown2020language,kirillov2023segment,radford2021learning,oquab2024dinov}. While some works utilize in-the-wild videos to extract intermediate representations, priors, or reward functions~\cite{chen2021learning,bahl2022human}, a key challenge remains: the lack of action labels. Patel et al.~\cite{patel2022learning} use a reconstruction pipeline to obtain hand-object trajectories as action labels for learning robot manipulation. SelfD~\cite{zhang2022selfd} argues BEV model is better than VO in pseudo-labeling, while we show VO works well with large-scale data. Recent works also rely on prompting closed VLMs to get action labels~\cite{wang2024vlm,wake2024gpt}. LeLaN~\cite{hirose2024lelan}, a concurrent work most relevant to ours, also relies on VLM prompting and uses pretrained navigation models to generate action labels. In contrast, we process in-the-wild video data differently by using only the noisy pseudolabels from off-the-shelf visual SLAM tools~\cite{teed2024deep}. This approach enables scalable and cost-effective action label generation for imitation learning in urban navigation tasks.

\section{Embodied Urban Navigation}

\subsection{Problem Formulation}

In this work, we address the problem of embodied urban navigation for agents in dynamic urban environments. The agent's task is to navigate from its current location to a specified target waypoint location, following a series of waypoints provided by a navigation tool. Formally, we define this as a point-goal navigation problem in real-world urban settings. At each time step $t$, the agent receives an RGB observation $o_t$, its current GPS location $p_t$, and a sub-goal waypoint $w_t$. The agent aims to learn a policy $\pi(a_t \mid o_{(t-k):t}, p_{(t-k):t}, w_t)$ that maps the past observations and positional information to an action $a_t$ from the action space $\mathcal{A}$, represented as a series of action waypoints in Euclidean space. Typically, we take $k=5$ and also predict actions for the next 5 time steps.

As accurate waypoints are easily available from navigation tools (e.g. Google Maps API), we focus our problem on the navigation between two consecutive waypoints. The agent treats the current sub-goal $w_t$ as the immediate target and is responsible for determining when it has reached this sub-goal based on its observations and positional data. Upon reaching $w_t$, the agent proceeds to the next waypoint in the sequence.

\subsection{Evaluation Metrics}\label{sec:metrics}
To enable analysis and assessment of different methods on the problem, our evaluation metrics are twofold. We use the ground truth trajectory in teleoperation-collected data to provide a thorough analysis of each method’s performance across various scenarios. Below, we discuss the metrics we used in offline data evaluation. For real-world deployment, we primarily use navigation success rate as the key metric for performance comparison.

\begin{figure}[t]
    \vspace{-4mm}
    \centering
    \includegraphics[width=1\linewidth]{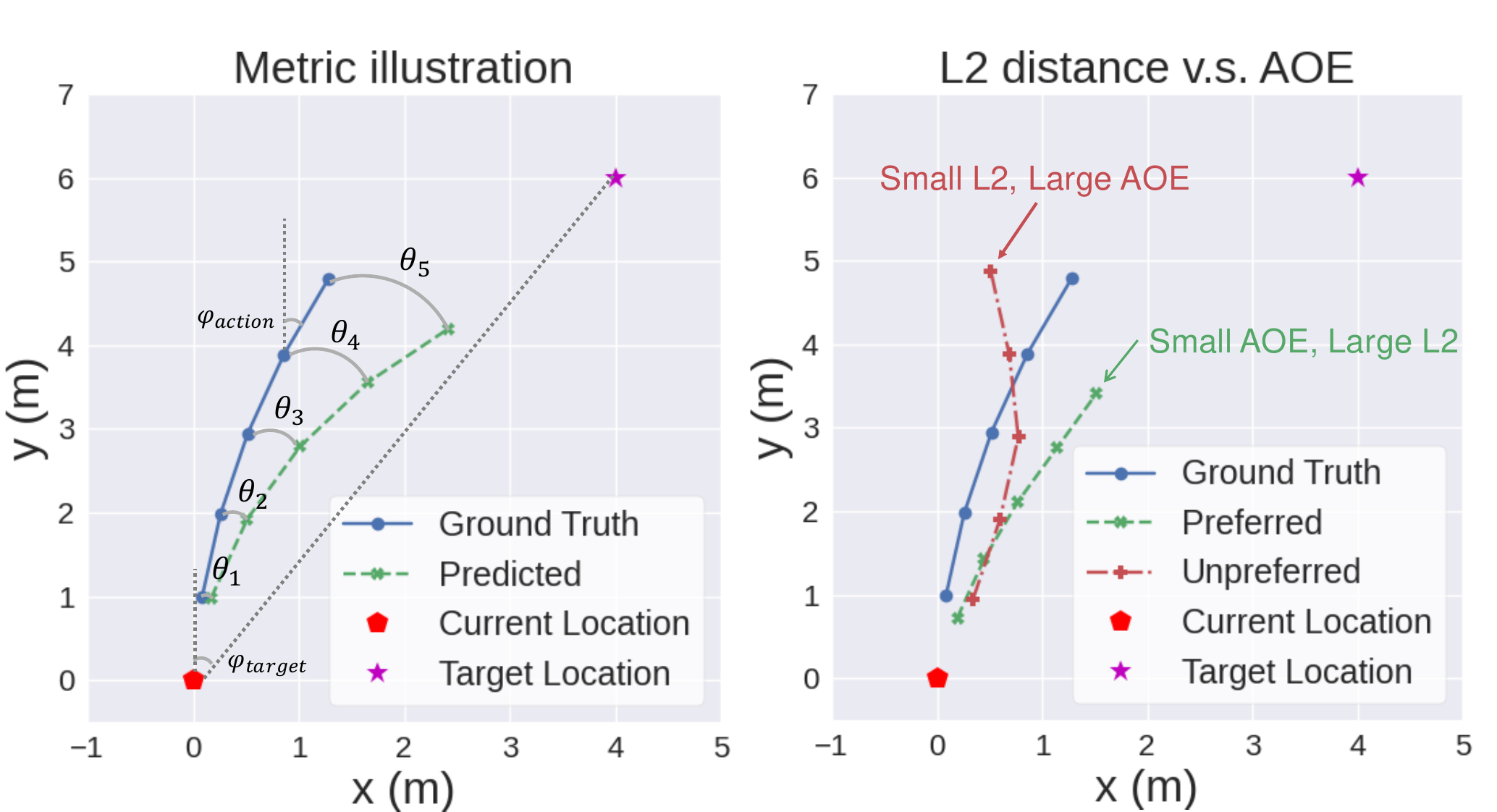}
    \vspace{-6mm}
    \caption{\textbf{Evaluation Metrics.} \textit{Left}: The orientation error is defined to be the angle between each predicted and ground truth action pair, labeled by $\theta_n$ in the figure. The action angle $\varphi_{\text{action}}$ and target angle $\varphi_{\text{target}}$ are defined with respect to the positive y-axis. \textit{Right}: Both green and red trajectories are predicted actions. The \textcolor{teal}{green} trajectory is the preferred one, having a large L2 distance but a small AOE. Vice versa for the \textcolor{sbred}{red} trajectory.}
    \label{fig:metric}
    \vspace{-4mm}
\end{figure}

\noindent\textbf{Average Orientation Error}. We use the error between each predicted action and the ground truth action as the main evaluation metric on the offline teleoperation data. While L2-distance in Euclidean space is a straightforward metric, we find it does not adequately capture the problem. As shown on the right side of \cref{fig:metric}, the red trajectory has a smaller L2 error compared to the green trajectory but moves in a direction that deviates from the target. This observation motivates us to come up with a new evaluation metric that better reflects the quality of the predicted actions. We define \textit{average orientation error} (AOE) as the angle difference between the predicted and ground truth actions, averaged across all samples at each step (left side of \cref{fig:metric}):
\begin{equation}
    \text{AOE}(k)=\frac{1}{n}\sum_i^n\theta_{i_k}=\frac{1}{n}\sum_i^n\arccos\frac{\langle\hat{a_{i_k}},a_{i_k}\rangle}{\Vert \hat{a_{i_k}}\Vert \Vert a_{i_k} \Vert},
\end{equation}
where $k$ is the index of predicted action and $n$ is the number of data. Additionally, we observe that taking the mean across $k$ predicted future time steps can lead to an underestimation due to steps with very small errors. Hence, we propose using the maximum average orientation error (MAOE) over all predicted actions as an overall estimation:
\begin{equation}
    \text{MAOE}=\frac{1}{n}\sum_i^n\max_k\theta_{i_k}.
\end{equation}
As orientation error might be noisy when action distance is small, we also use L2-distance as a complimentary evaluation metric.


\noindent\textbf{Critical Scenarios}. Not all time steps are equally critical in a trajectory. For example, errors at road crossings or turning points are more likely to result in navigation failure compared to mistakes during forward movement in open areas. We identify several key scenarios that are most critical during navigation. AOE and MAOE are calculated separately for each scenario, and we use the scenario mean rather than the sample mean for a more comprehensive evaluation. Detailed definitions of each scenario are provided in \cref{sec:experiment}.

\subsection{Learning from In-the-Wild Videos}
\label{sec:wild-video}
\begin{figure}[t]
    \centering
    \vspace{-5mm}
    \includegraphics[width=1\linewidth]{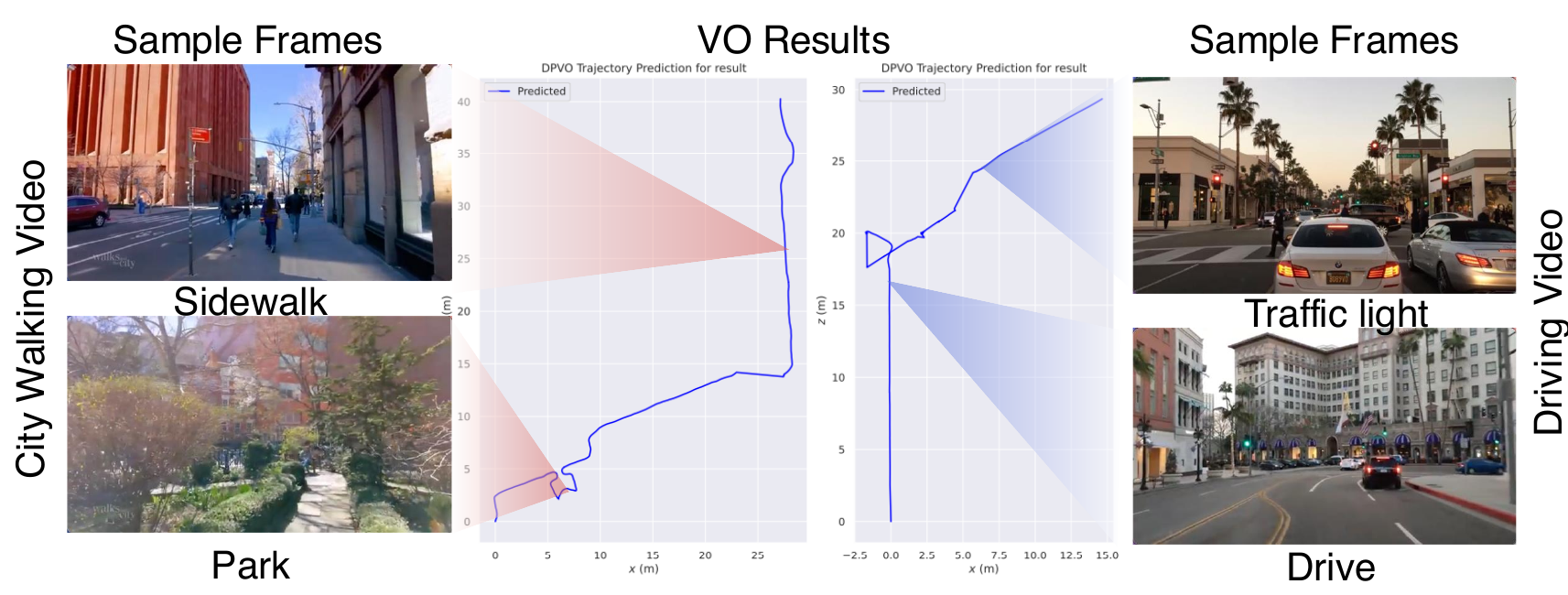}
    \caption{\textbf{Data Sample and Visual Odometry (VO) Result.} Our internet-source training data includes both walking and driving videos. These videos cover various scenarios in the urban environment. The VO tool gives noisy trajectories globally, but trustworthy local relative pose within a short time period.}
    \label{fig:dpvo}
    \vspace{-4mm}
\end{figure}

\textbf{City Walking Videos}. To facilitate large-scale imitation learning for human-like urban navigation, we utilize city walking videos sourced from the internet. Walking in urban environments is intuitive for humans, resulting in an abundance of first-person view videos available online. These videos naturally capture the complexities of navigating dynamic urban landscapes, including interactions with pedestrians, adherence to traffic signals, and maneuvering around various obstacles. Motivated by the extensive availability of such data and its inherent alignment with our navigation objectives, we curate a dataset comprising over 2000 hours of city walking videos, encompassing diverse weather and lighting conditions. These videos exhibit consistent motion patterns characteristic of walking, providing continuous motion cues that can directly serve as action labels.

\noindent\textbf{Action Labels from Videos}. To generate actionable labels for imitation learning, we employ VO~\cite{engel2017direct,mur2015orb,teed2021droid,teed2024deep} tools to extract trajectory poses from the collected city walking videos. Specifically, we utilize the state-of-the-art DPVO~\cite{teed2024deep} while any other VO method should also work. The resulting VO trajectories are illustrated in \cref{fig:dpvo}. However, using VO poses presents two primary challenges. First, global trajectory accuracy is compromised by accumulation errors inherent in VO systems. Nevertheless, since our model focuses on predicting a limited sequence of actions (5 past and 5 future), we mitigate this issue by relying solely on relative poses within short temporal windows. Second, VO methods suffer from scale ambiguity, leading to inconsistencies in trajectory lengths across different data. For instance, as shown in \cref{fig:dpvo}, a driving video may exhibit a longer actual trajectory compared to a walking video, yet the DPVO output incorrectly reflects a shorter trajectory. To address this, we normalize each action by the average step length within its trajectory, following the approach in previous work~\cite{shah2023gnm}. This normalization resolves both scale ambiguity and inconsistencies in step lengths between different types of embodiments all at once. This enables the learning of policies in a consistent, abstracted action space. During deployment, the predicted actions are denormalized using the specific step length of the target robot, ensuring appropriate movement execution in real-world settings.

\noindent\textbf{Generalizability and Scalability.} Our data processing pipeline is both generalizable and scalable. It can be applied to any video sources beyond city walking, as long as the videos are captured with egomotion, such as driving videos, which are also abundantly available online. This flexibility enables the training of more generalizable navigation policies that can be zero-shot applied or fine-tuned for cross-domain and cross-embodiment tasks, such as transitioning from pedestrian to vehicular navigation.  Furthermore, our method offers significant scalability advantages. Unlike pipelines that rely on prompting large language models for action label generation~\cite{wang2024vlm,hirose2024lelan}, our visual odometry-based approach can be easily parallelized. Processing 2000 hours of video data with our pipeline requires negligible wall-clock time, making it both feasible and cost-efficient to scale up the training dataset.

\subsection{Pipeline and Training}

\textbf{Pipeline Architecture}. \Cref{fig:pipeline} shows our pipeline. It builds upon previous works~\cite{shah2023vint,sridhar2024nomad}. At the core of our model is a transformer that processes a temporal sequence of input tokens, including $k$ image features and one coordinate embeddings. Specifically, the coordinate input is stacked by $k$ past positions and one single target position, and outputs a sequence of the same length. We use an action head and an arrival prediction head to decode the output of the transformer. We utilize DINOv2~\cite{oquab2024dinov} representation and freeze the backbone during training.

\noindent\textbf{Feature Hallucination}. Inspired by previous works on feature learning~\cite{he2023metric,assran2023self,bardes2024revisiting}, we use feature hallucination as an auxiliary loss during training. Specifically, we compute the MSE loss between the output image tokens and that directly extracted from future frames. We hope this will guide the transformer to predict more informative future tokens that best mimic future frames, which in turn helps the two MLP heads predict the final action and arrival status.

\noindent\textbf{Loss Functions}. As mentioned in \cref{sec:metrics}, L1 or L2 loss can be less effective than orientation error in evaluating the quality of the predicted actions. We also employ orientation error to supervise the training. We define the orientation loss to be the negative cosine similarity between the predicted actions and the groun truth actions:
\begin{equation}
    \mathcal{L}_{\text{ori}}=-\frac{1}{k}\sum_{i=1}^k\frac{\langle\hat{a_{i}},a_{i}\rangle}{\Vert \hat{a_{i}}\Vert \Vert a_{i} \Vert}.
\end{equation}
Together with the L1 loss for predicted action and the BCE loss for predicted arriving status, the final loss function is a weight sum of four individual losses:
\begin{equation}
    \mathcal{L}=\omega_\text{l1}\mathcal{L}_{l1}+\omega_\text{ori}\mathcal{L}_{ori}+\omega_\text{arr}\mathcal{L}_{arr}+\omega_\text{feat}\mathcal{L}_{feat}.
\end{equation}
We choose each weight so that each loss items are within the same magnitude.

\section{Experiment}\label{sec:experiment}
\begin{table*}[t]
    \centering
    \caption{\textbf{Benchmark on Offline Data}. We evaluate three metrics in each critical scenario for all methods. Percentages under scenarios indicate their data proportions. The ``Mean" column shows scenario means averaged over six scenarios; ``All" shows sample means over all data samples. We highlight the best performance in \textbf{bold} and the second-best performance with \underline{underline}.}\label{tab:benchmark}
    \resizebox{0.8\linewidth}{!}{
    \begin{tabular}{ll|cccccccc}
    \toprule
        \multirow{2}{*}{\textbf{Method}} & \multirow{2}{*}{\textbf{Metric}} & \multirow{2}{*}{\textbf{Mean}} & \textbf{Turn} & \textbf{Crossing} & \textbf{Detour} & \textbf{Proximity} & \textbf{Crowd} & \textbf{Other} & \textbf{All} \\ 
        & & & 8\% & 12\% & 12\% & 6\% & 7\% & 55\% & 100\% \\ 
        \midrule
        \multirow{3}{5em}{GNM~\cite{shah2023gnm} (fine-tuned)} 
         & $\downarrow$ L2 (m) & \underline{1.22} & 2.36 & 1.36 & 1.42 & \underline{0.88} & \underline{0.76} & \textbf{0.55} & \underline{0.74}\\
         & $\downarrow$ MAOE ($^{\circ}$) & \underline{16.2} & 31.1 & \underline{14.8} & \textbf{12.5} & \underline{14.7} & \underline{12.8} & \underline{11.0} & \underline{12.1}\\
         & $\uparrow$ Arrival (\%) & 68.6 & 66.4 & 69.7 & 66.4 & 69.0 & 69.2 & 70.7 & 70.0 \\ 
         \midrule
        \multirow{3}{5em}{ViNT~\cite{shah2023vint} (fine-tuned)} 
         & $\downarrow$ L2 (m) & 1.30 & 1.91 & 1.13 & \textbf{1.14} & \textbf{0.77} & \textbf{0.66} & \underline{0.57} & \textbf{0.70}\\
         & $\downarrow$ MAOE ($^{\circ}$) & 16.5 & 31.1 & 15.4 & \underline{12.9} & 14.8 & 13.3 & 11.6 & 12.6\\
         & $\uparrow$ Arrival (\%) & 70.5 & \textbf{71.2} & 68.7 & 70.7 & 73.0 & 68.6 & 71.0 & 70.7 \\ 
         \midrule
        \multirow{3}{5em}{NoMaD~\cite{sridhar2024nomad} (fine-tuned)} 
         & $\downarrow$ L2 (m) & 1.39 & 2.49 & 1.56 & 1.55 & 1.06 & 0.95 & 0.76 & \underline{0.74}\\
         & $\downarrow$ MAOE ($^{\circ}$) & 19.1 & 35.1 & 18.5 & 15.6 & 18.1 & 14.3 & 12.8 & \underline{12.1}\\
         & $\uparrow$ Arrival (\%) & 68.6 & 66.4 & 69.7 & 66.4 & 69.0 & 69.3 & 70.7 & 70.0 \\ 
         \midrule
        \multirow{3}{5em}{\textbf{Ours} (zero-shot)} 
         & $\downarrow$ L2 (m) & \cc 1.34 & \cc \underline{1.30} & \cc \underline{1.09} & \cc 1.33 & \cc 1.44 & \cc 1.48 & \cc 1.39 & \cc 1.38\\ 
         & $\downarrow$ MAOE ($^{\circ}$) & \cc 16.5 & \cc \underline{26.8} & \cc 15.5 & \cc 16.3 & \cc 16.3 & \cc 13.1 & \cc 11.3 & \cc 12.7\\ 
         & $\uparrow$ Arrival (\%) & \cc \underline{79.1} & \cc \underline{69.3} & \cc \underline{71.4} & \cc \textbf{78.8} & \cc \underline{84.1} & \cc \underline{84.5} & \cc \underline{86.4} & \cc \underline{84.1} \\ 
         \midrule
        \multirow{3}{5em}{\textbf{Ours} (fine-tuned)} 
         & $\downarrow$ L2 (m) & \cc \textbf{1.11} & \cc \textbf{1.27} & \cc \textbf{1.00} & \cc \underline{1.15} & \cc 1.06 & \cc 1.12 & \cc 1.06 & \cc 1.07\\ 
         & $\downarrow$ MAOE ($^{\circ}$) & \cc \textbf{15.2} & \cc \textbf{26.6} & \cc \textbf{14.1} \cc & \cc 13.9 & \cc \textbf{14.3} & \cc \textbf{12.0} & \cc \textbf{10.4} & \cc \textbf{11.5}\\ 
         & $\uparrow$ Arrival (\%) & \cc \textbf{81.8} & \cc 68.9 & \cc \textbf{75.3} & \cc \underline{78.5} & \cc \textbf{90.6} & \cc \textbf{87.5} & \cc \textbf{90.2} & \cc \textbf{87.8}\\
    \bottomrule
    \end{tabular}
    \vspace{-4mm}
}
\end{table*}

\subsection{Setup}
\textbf{Baselines}. We compare our model with outdoor navigation models closely related to our setup, including GNM~\cite{shah2023gnm}, ViNT~\cite{shah2023vint}, and NoMaD~\cite{sridhar2024nomad}. Although originally developed for image-goal navigation, we tested them using goal-images from our collected data. Additionally, ViNT as a foundation model for visual navigation supports fine-tuning, which we also evaluated. We acknowledge CoNVOI~\cite{sathyamoorthy2024convoi} as a recent work with a similar setup, but could not test it due to the lack of open source code and prompt.  We also experimented with VLM-based approaches similar to CoNVOI and defer the results to the appendix.

\noindent\textbf{Data}. We collected expert data via teleoperation for fine-tuning and offline testing. We collect data via a Unitree Go1 quadruped equipped with a Livox Mid-360 LiDAR and a webcam for RGB observations. We employed the LiDAR-SLAM method~\cite{xu2021fast} to obtain the robot's poses as ground truth actions. Additionally, we retrieved location data from a smartphone using a webpage. We reduce noise from the quadruped's vibrations by holding the phone~\cite{sorokin2022learning}. In total, we gathered 15 hours of teleoperation data across various areas in New York City, allocating 6 hours for fine-tuning and 9 hours for testing.

\noindent\textbf{Critical Scenarios} As discussed in \cref{sec:metrics}, we identify several critical scenarios in our dataset that we want to emphasize in our evaluation. We analyze the ground truth trajectory and run object detection~\cite{he2017mask} to aid us in standardizing the definition. We define these scenarios as follows:
\begin{itemize}
    \item \textit{Turn}: When the ground truth action changes direction significantly. Defined when $\varphi_\text{action}>20^\circ$ (see \cref{fig:metric}).
    \item \textit{Crossing}: When the agent is at a road crossing. Defined if any traffic light is detected with score $>0.5$.
    \item \textit{Detour}: When the action angle deviates from the target angle. Defined when $|\varphi_\text{action}-\varphi_\text{target}|>45^\circ$ (see \cref{fig:metric}).
    \item \textit{Proximity}: When any person is close to the agent. Defined when the largest bounding box of any detected person is larger than 25\% of the image area.
    \item \textit{Crowd}: When the agent is surrounded by a crowd of people. Defined when $\geq 5$ people are detected.
\end{itemize}
We define these scenarios in a way that they are not mutually exclusive. A single data sample can belong to multiple scenarios. Although these scenarios account for less than half of the data, we argue they are the most significant factors contributing to successful urban navigation.

\subsection{Performance Benchmarking}

In this subsection, we aim to answer the question \textbf{Q1}: Can our CityWalker model navigate successfully in complex urban environments? To answer this, we benchmark our model and the baselines both with offline data and in real-world navigation deployment.

\noindent\textbf{Overall Benchmark}. \Cref{tab:benchmark} shows the benchmark in different critical scenarios. Overall, our fine-tuned model has superior performance on all metrics in all scenarios except ``detour''. We think the less competent performance in the detour scenario is due to the small portion of such data in our video training data. It can be verified by the fact that the fine-tuning performance gain is most significant in this scenario. It is worth noting that our zero-shot model has similar or even better performance compared to the fine-tuned baselines. This demonstrates the power of large-scale pretraining, especially considering the domain and modality gap from human walking videos.

\noindent\textbf{Real-world Depolyment}. We deploy our model and the baseline methods on the same Unitree Go1 quadruped for real-world navigation, in previously unseen urban environments. We adapt the PD controller provided in \cite{shah2023vint} to provide velocity command to the quadruped. We choose area with relatively constant density of dynamics such as pedestrian and traffic to ensure the consistency among different trials. In each trial, the target location is generally 50-100m far away from the starting location. We categorize the real-world test into \textit{forward}, \textit{left turn}, and \textit{right turn} cases to provide case-by-case analysis. We conduct 8-14 trails for each cases. Each trial is marked success when the model predicts arrival within a 5m distance to the target location. All human interruptions due to potential collision or timeout are treated as failure cases.

\begin{table}[t]
    \caption{\textbf{Real-World Navigation}. The table shows the success rate of real-world experiments in different scenarios. $*$ and $\dagger$ indicate zero-shot inference and fine-tuned models respectively.}
    \label{tab:realworld}
    \vspace{-2mm}
    \centering
    \resizebox{0.9\linewidth}{!}{
    \begin{tabular}{l|ccccc}
    \toprule
        \textbf{Method} & \bf All & \bf Forward & \bf Left turn & \bf Right turn \\ \midrule
        ViNT$^*$\cite{shah2023vint} & 37.7 & 62.5 & 0.0 & 50.0 \\
        ViNT$^\dagger$\cite{shah2023vint} & 57.1 & 100.0 & 25.0 & 25.0 \\
        NoMad$^*$\cite{sridhar2024nomad} & 42.9 & 75.0 & 16.7 & 28.6 \\
        \midrule
        \cc \textbf{Ours}$^\dagger$ & \cc \bf 77.3 & \cc \bf 100.0 & \cc \bf 62.5 & \cc \bf 66.7 \\
        \bottomrule
    \end{tabular}
    
    }
    \vspace{-4mm}
\end{table}

\begin{figure*}[t]
    \centering
    \includegraphics[width=0.95\linewidth]{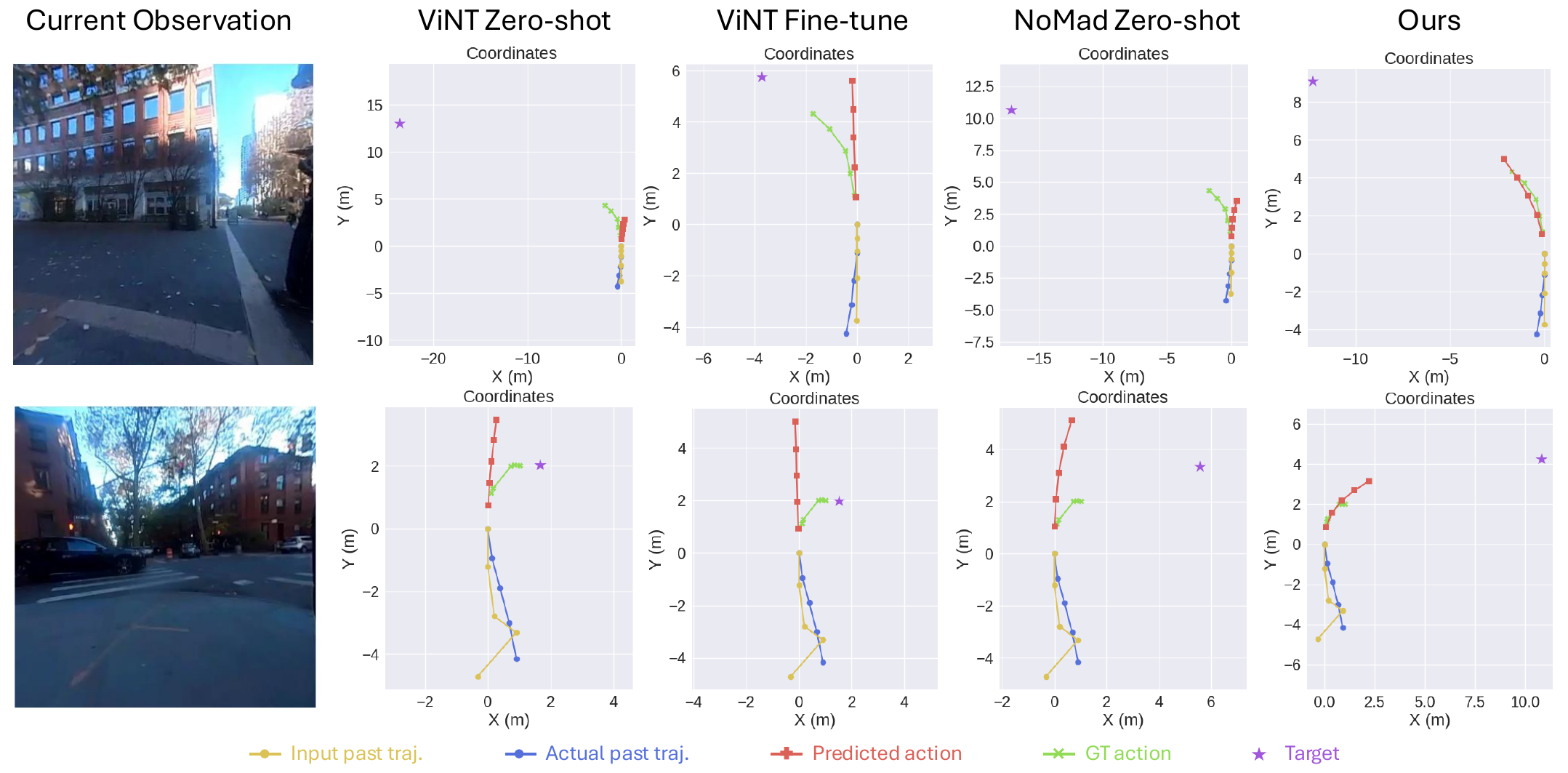}
    \vspace{-2mm}
    \caption{\textbf{Qualitative Results}. Left image shows current observations of two samples. Right plots displays input trajectory, ground truth actions, and predicted actions in the current coordinate system with the agent at the origin.}
    \label{fig:qualitative}
    \vspace{-3mm}
\end{figure*}

\noindent \Cref{tab:realworld} shows the success rate on all real-world test cases. Our model achieves the highest success rate among all cases with a significant gap to the second-best performing fine-tuned ViNT. As the baselines are trained with navigation data mostly in suburban or off-road environments~\cite{shah2023gnm,hirose2023sacson}, they are limited in handling complex maneuvers, making them unsuitable for urban navigation tasks, as shown in \cref{fig:qualitative}. Our model’s robust performance in both forward and turns highlights its ability to manage dynamic and varied urban scenarios effectively, allowing it to be more reliably used in real-world environments where frequent and precise directional changes are essential. The substantial improvement not only increases the overall navigation success but also extends the applicability of our model to more demanding and realistic urban settings, positioning it as a superior solution for embodied agents in dynamic environments.

\begin{figure}[t]
    \centering
    \includegraphics[width=0.95\linewidth]{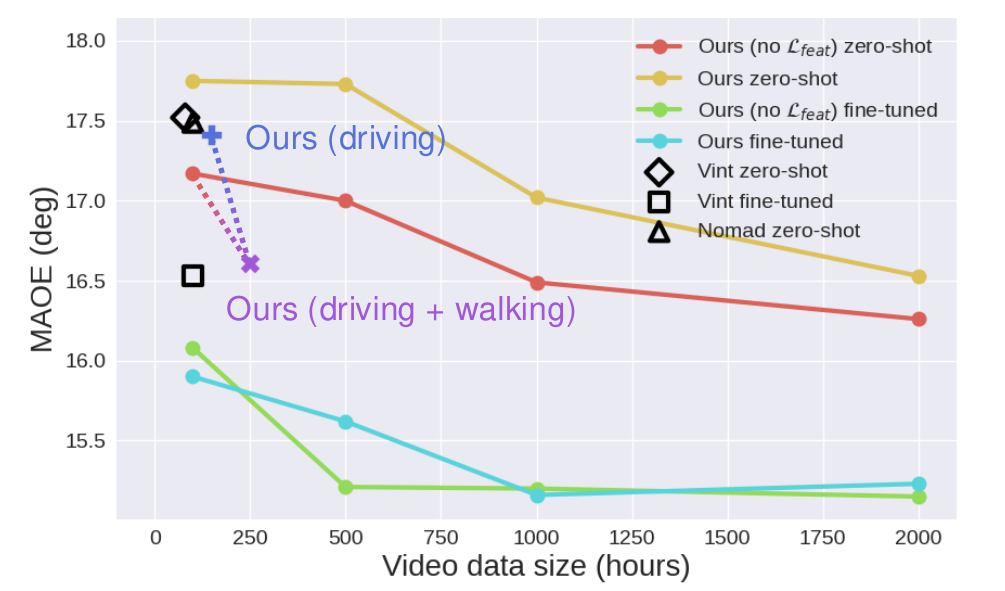}
    \vspace{-2mm}
    \caption{\textbf{Performance and Data Size}. We show the model performance evaluated by MAOE with respect to the size of the training data measured by video length in hours. We also show the zero-shot performance of our model trained with only driving videos and mixed driving and walking videos} 
    \label{fig:scale}
    \vspace{-2mm}
\end{figure}

\subsection{Power of Data Scaling}
Our model is significantly boosted by the large-scale training data. We ask \textbf{Q2}: To what extent does increasing the size of the training dataset improve the performance of our urban navigation model?

\noindent\textbf{Performance Gain and Data Scale}. \Cref{fig:scale} shows the model performance when trained with different data sizes. The plot demonstrates that our model's zero-shot performance improves significantly as we increase the amount of training data. Notably, when trained with more than 1000 hours of videos, our zero-shot model (\textcolor{sbred}{red line}) achieves a better performance than the fine-tuned ViNT model ($\square$). This indicates that large-scale training on diverse video data can even surpass the performance of models that are fine-tuned with limited expert data, highlighting the potential of leveraging web-scale datasets for embodied navigation.

\noindent\textbf{Domain and Embodiment Gap}. An unexpected observation is that our model \textit{without} the feature hallucination loss outperforms the one \textit{with} it in zero-shot inference. We hypothesize this phenomenon mainly due the domain and embodiment gap between human walking videos and quadruped navigation policies. The feature hallucination loss aims to guide the transformer model to produce output tokens that mimic future observations, essentially encouraging the model to anticipate future visual inputs. This enforcement might inadvertently hinder performance by focusing on predicting future observations based on human motion. The model may learn representations that are less applicable or even misleading for quadruped navigation. Consequently, the feature hallucination loss could introduce a bias that detracts from the model's ability to make accurate navigation decisions in the robot's operational domain. 

\begin{figure}[t]
    \centering
    \includegraphics[width=0.9\linewidth]{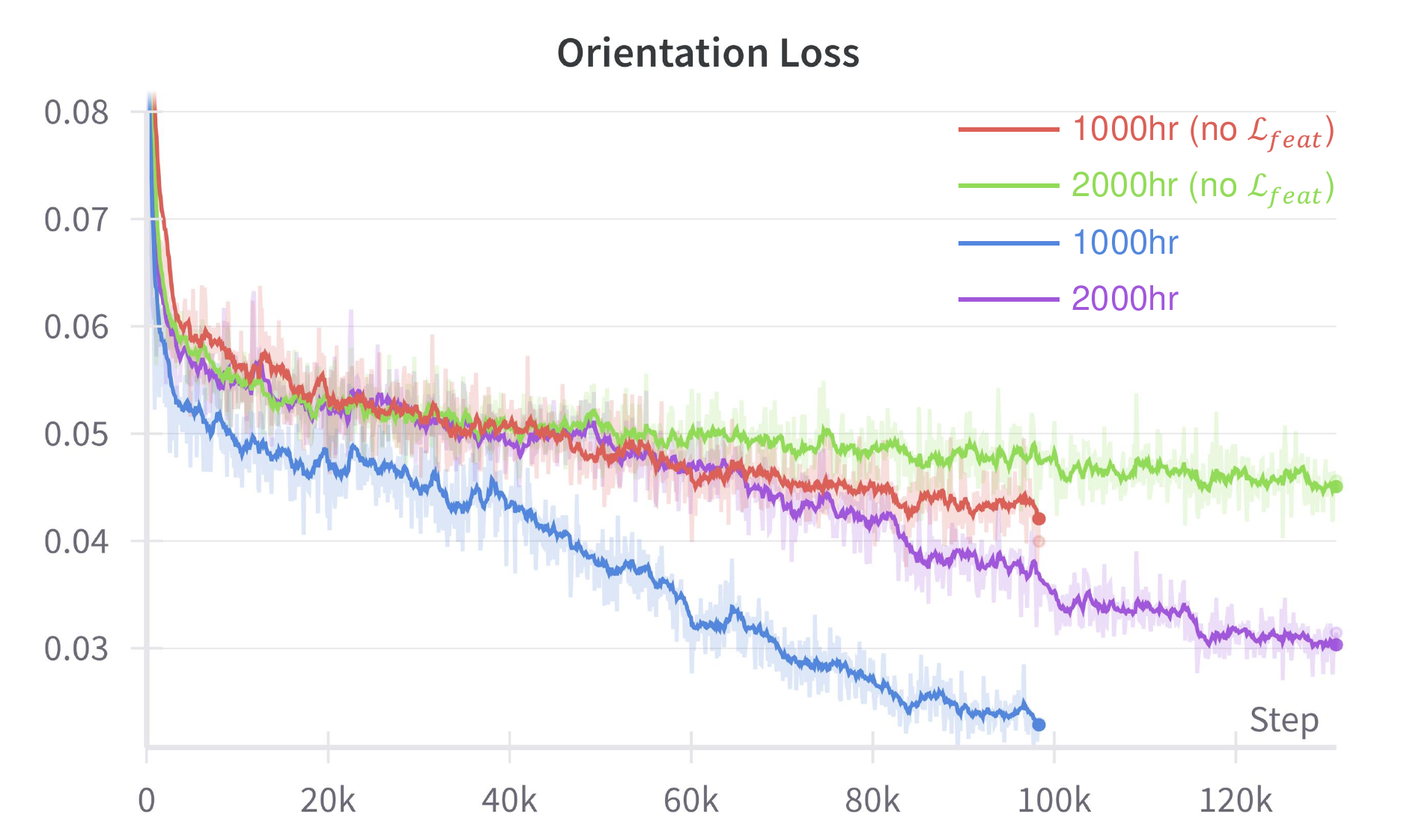}
    \caption{\textbf{Improvement from Feature Hallucination}. The plot shows the orientation loss for different amounts of training data and loss setups. For all four setups, we trian with $\omega_{\text{l1}}=1.0,, \omega_{\text{arr}}=1.0,and, and \omega_\text{ori}=5.0$. The unproportional training steps are due to different batch sizes used for different data sizes. All models are trained for 10 epochs.}
    \label{fig:loss}
    \vspace{-2mm}
\end{figure}

\noindent\textbf{Feature Hallucination Loss}. Fortunately, as shown in \cref{tab:benchmark}, this issue no longer exists after fine-tuning. We also show the benefit of feature hallucination loss from another perspective. \Cref{fig:loss} shows two training loss curve for different data size and training setups with and without feature hallucination loss. We observe that although feature hallucination loss should increase the absolute value of the total loss, the training with feature hallucination loss lead to a lower total loss for 2000-hour training and a lower orientation loss in both cases. The training with feature hallucination loss also shows a steeper decreasing trend, showing a potential of having even lower error with more training steps. Therefore, this is another proof of the effectiveness of our proposed feature hallucination loss.

\noindent\textbf{Driving Videos}. As discussed in \cref{sec:wild-video}, our data processing pipeline also enables the use of abundant online driving videos. We evaluate models trained exclusively on driving videos and on a combination of walking and driving videos, results shown in \cref{fig:scale}. We can observe that the model trained only with driving data (\textbf{\textcolor{sbblue}{+}}) has a similar performance to the zero-shot baselines. As a result, we can think of driving as a form of navigation from a different domain and embodiment. However, combining driving and walking data yields significant performance gains ({\textcolor{sbpurple}{$\mathbf{x}$}}). With just 250 hours of mixed data, the model nearly matches the performance of models trained on 1000 hours of walking data alone. This signifies the generalizability and robustness provided by cross-domain cross-embodiment data.

\subsection{Pipeline component analysis}
To provide a more in-depth analysis of our model and pipeline, we ask \textbf{Q3}: How do the individual components of our pipeline enhance the performance and reliability of our model for real-world urban navigation?

\noindent\textbf{AOE at Different Time Steps}. In \cref{tab:aoe}, we study the AOE at different time steps, i.e. AOE$(k)$ for $k$ from $1$ to $5$. ViNT exhibits increasing orientation errors as the prediction steps progress, indicating a degradation in accuracy over time. In contrast, our model maintains relatively consistent AOE from step 2 to step 5, which likely contributes to its superior performance in real-world navigation deployments with stable action prediction across multiple future steps. All methods experience higher errors at the first step. We attribute it to the proximity of the initial action to the origin, as minor linear deviations result in substantial angular error. It is worth noting that the MAOE is significantly larger than any individual AOE from steps 1 to 5, suggesting that the peak errors occur at varying time steps across different samples. This variability suggests the potential benefit of taking softmax for orientation errors instead of simple averaging, as it could better capture the distribution of peak errors.

\begin{table}[t]
    \caption{\textbf{Ablation Study}. We ablate on the contribution of orientation loss, feature hallucination, and fine-tuning with expert data. We show the MAOE($^{\circ}$) in scenario-mean. All results are trained with 1000-hour walking data.}
    \label{tab:ablation}
    \centering
    \resizebox{0.8\linewidth}{!}{
    \begin{tabular}{ccc|c}
    \toprule
        \multicolumn{3}{c|}{\bf Training Components} & \textbf{MAOE} \\ 
        Ori. Loss & Feature Hall. & Fine-tuning & Mean \\ \midrule
        
         & & & 17.03 \\
        
        \checkmark & & & 17.00 \\
        
        \checkmark & \checkmark & & 17.02 \\
         
         &  & \checkmark & 15.23 \\
        
        \checkmark & & \checkmark & \underline{15.21} \\
        
        \checkmark & \checkmark & \checkmark & \textbf{15.16} \\
        \bottomrule
    \end{tabular}
    }
\end{table}

\begin{table}[t]
    \caption{\textbf{AOE at Different Time Steps}. The table shows MAOE and different AOE of scenario-mean, all in degrees ($^\circ$). $*$ and $\dagger$ indicates zero-shot inference and fine-tuned models respectively.}
    \label{tab:aoe}
    \centering
    \resizebox{\linewidth}{!}{
    \begin{tabular}{l|cccccc}
    \toprule
        \textbf{Method} & \bf AOE(1) & \bf AOE(2) & \bf AOE(3) & \bf AOE(4) & \bf AOE(5) & \bf MAOE \\ \midrule
        ViNT$^*$\cite{shah2023vint} & \bf 10.57 & 7.93 & 9.09 & 10.21 & 11.45 & 17.52 \\
        ViNT$^\dagger$\cite{shah2023vint} & 11.24 & 7.96 & 8.40 & 8.93 & 9.87 & 16.53 \\
        \midrule
        \cc \textbf{Ours}$^*$ & \cc 12.91 & \cc 8.92 & \cc 8.64 & \cc 8.46 & \cc 8.43 & \cc 16.54 \\
        \cc \textbf{Ours}$^\dagger$ & \cc 11.44 & \cc \bf 7.77 & \cc \bf 7.68 & \cc \bf 7.75 & \cc \bf 7.97 & \cc \bf 15.23\\
        \bottomrule
    \end{tabular}
    }
    \vspace{-2mm}
\end{table}

\noindent\textbf{Ablation Study}. \Cref{tab:ablation} examines the individual contributions of different pipeline components to our model's performance. A clear improvement in MAOE when fine-tuning with expert data, demonstrating the effectiveness of using in-domain expert information. The improvement from orientation loss and feature hallucination loss is marginal, within error range of our limited testing data. We defer more detailed ablation study to the appendix.

\section{Conclusion}

\noindent\textbf{Summary}. In this work, we addressed embodied urban navigation by training agents with thousands of hours of in-the-wild city walking videos, introducing a scalable pipeline for imitation learning from diverse walking videos. Our experiments show that large-scale, diverse training data significantly enhances navigation performance, surpassing existing methods and demonstrating the benefits of data scaling. Our findings highlight the potential of leveraging abundant online videos to develop robust navigation policies for embodied agents in dynamic urban environments.

\noindent\textbf{Limitation and Future Work}. During real-world deployment, our current system implementation is sensitive to potential large location noises read naively from iPhone location services, which could be improved with better GPS hardware. Future work could also enhance the robustness of the model to such location noises.

\section*{Acknowledgment}

The work was supported by NSF grants 2238968, 2121391, 2322242 and 2345139; and in part through the NYU IT High Performance Computing resources, services, and staff expertise. We also thank Xingyu Liu and Zixuan Hu for their help in data collection.

{
    \small
    \bibliographystyle{unsrt}
    \bibliography{main}
}

\clearpage
\clearpage
\setcounter{page}{1}
\renewcommand{\thesection}{\Alph{section}}
\renewcommand{\thefigure}{\Roman{figure}}
\renewcommand{\thetable}{\Roman{table}}

\setcounter{section}{0}
\setcounter{figure}{0}
\setcounter{table}{0}

\section*{Appendix}


\section{Details on Data, Model, and Training}
\begin{figure}[t]
    \centering
    \includegraphics[width=0.65\linewidth]{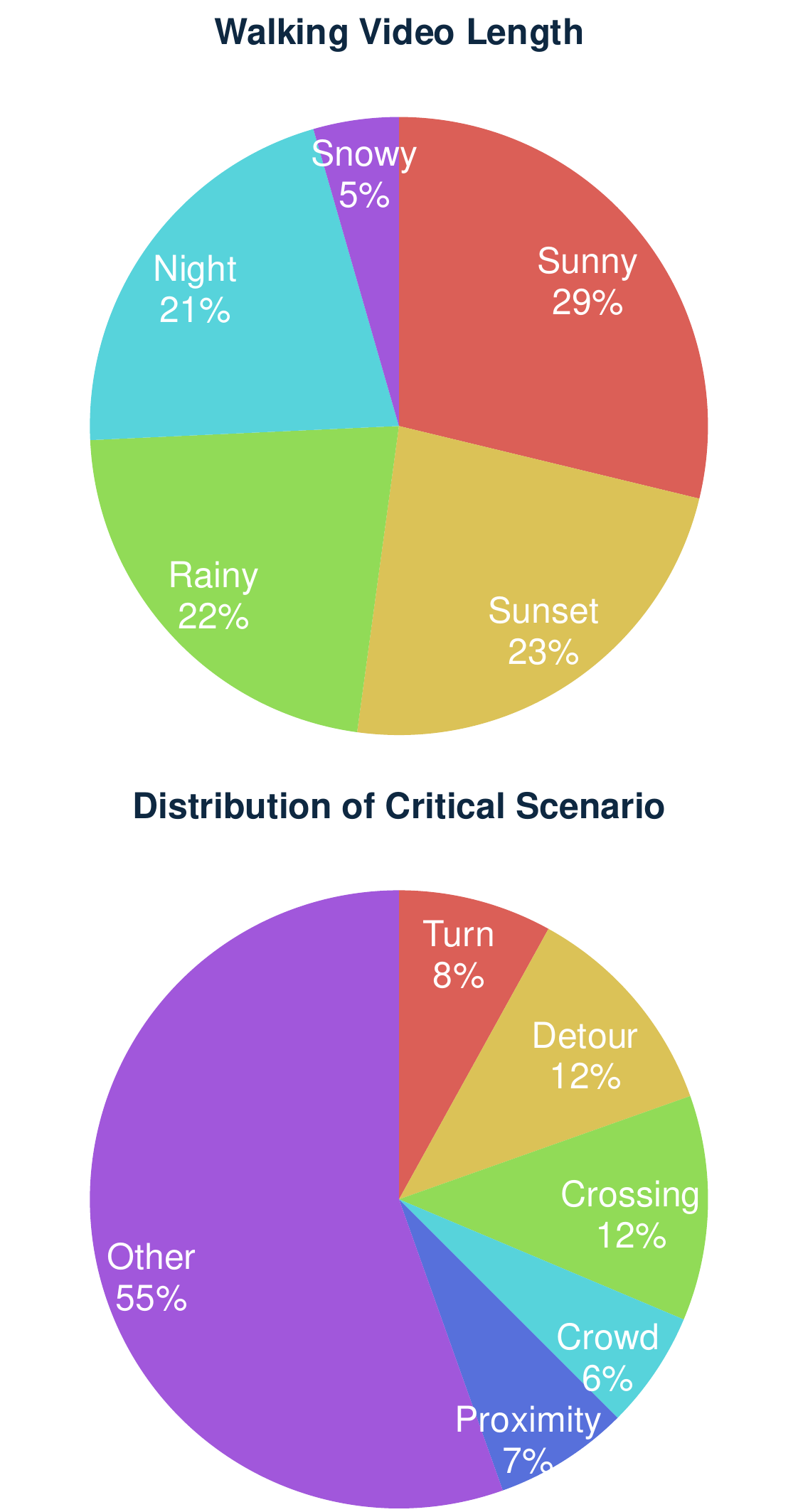}
    \caption{\textbf{Data distribution}. \textit{Top}: The distribution of different weather and lightning conditions in our video training data. \textit{Bottom}: The distribution of different critical scenarios in our collected data.}
    \label{fig:distribution}
    \vspace{-4mm}
\end{figure}

\textbf{City Walking Videos}. We source our training video data mainly from the city walking\footnote{\href{https://www.youtube.com/@WALKS_and_the_CITY/playlists}{https://www.youtube.com/@WALKS\_and\_the\_CITY/playlists}} and driving\footnote{\href{https://www.youtube.com/@jutah/playlists}{https://www.youtube.com/@jutah/playlists}} playlists on YouTube. The full sourced videos have a total length of 2522 hours. We use 2000 hours of them for training. These videos cover different weather and lighting conditions. \Cref{fig:distribution} shows a detailed distribution of each condition.

The lower part of \cref{fig:distribution} illustrates the proportion of each critical scenario in our offline expert data based on our definitions. We observe that the union of critical scenarios accounts for less than half of the dataset. However, these scenarios contribute most to the success rate in real-world experiments. This highlights the need for future work to enhance model performance in these critical areas.

\noindent \textbf{Hyperparameters for Model and Training}.
For model and training hyperparameters, we largely follow previous work~\cite{shah2023vint} and adapt some parameters to our case, as shown in \cref{tab:hyperparameter}. Note that DINOv2~\cite{oquab2024dinov} uses ViT~\cite{dosovitskiy2020vit} so it can adapt to any input resolution as long as it is divisible by the patch size. Therefore, we center-crop the $360 \times 640$ city walking videos to $350 \times 630$, and the $400 \times 400$ teleoperation video to $392 \times 392$ to keep the aspect ratio and as much visual content as possible.

\section{More Quantitative Results}

\textbf{Full Ablation Study}. In \cref{tab:full-ablation}, we provide an extended ablation study, including all the critical scenarios. We can observe that the addition of orientation loss and feature hallucination loss does not result in significant performance improvements. This lack of noticeable enhancement can be attributed to several factors, including the limited size of our training data (1000 hours) and the constrained nature of our test dataset, which is prone to substantial noise in the evaluation results. Consequently, we consider errors beyond the decimal point to be negligible.

Another interesting observation is the decline in performance within the Turn scenario following fine-tuning. We attribute this performance drop to the disproportionate representation of Turn scenarios in our fine-tuning data (8\%) compared to the original video data (32\%), leading to insufficient training examples for effectively handling turns.

\noindent \textbf{VLM Performance}. In \cref{tab:gpt-result}, we present the performance of the VLM (GPT-4o~\cite{achiam2023gpt}) on our urban navigation tasks. Our results indicate that GPT-4o struggles to generate reasonable navigation actions off-the-shelf via prompting. However, it performs reasonably in predicting the arrival status, likely because this sub-task is inherently more straightforward given the input of past and target locations.

\begin{table*}[t]
    \caption{\textbf{Full Ablation Study}. Here we provide a extended ablation study in supplementary \cref{tab:ablation}. The result is evaluated for all scenarios.}
    \label{tab:full-ablation}
    \centering
    \resizebox{0.85\linewidth}{!}{
    \begin{tabular}{ccc|cccccccc}
    \toprule
        \multicolumn{3}{c|}{Training Components} & \multirow{2}{*}{\textbf{Mean}} & \multirow{2}{*}{\textbf{Turn}} & \multirow{2}{*}{\textbf{Crossing}} & \multirow{2}{*}{\textbf{Detour}} & \multirow{2}{*}{\textbf{Proximity}} & \multirow{2}{*}{\textbf{Crowd}} & \multirow{2}{*}{\textbf{Other}} & \multirow{2}{*}{\textbf{All}}\\ 
        Ori. Loss & Feature Hall. & Fine-tuning & & & & & & & & \\ \midrule
        
         & & & 17.03 & \textbf{27.09} & 16.25 & 16.72 & 16.99 & 13.28 & 11.88 & 13.16 \\
        
        \checkmark & & & 17.00 & \underline{27.14} & 16.40 & 16.43 & 16.74 & 13.19 & 12.12 & 13.32 \\
        
        \checkmark & \checkmark & & 17.02 & 27.17 & 15.92 & 16.51 & 17.19 & 13.23 & 12.10 & 13.32  \\
         
         &  & \checkmark & 15.23 & 28.94 & \textbf{13.90} & \underline{13.14} & \underline{14.39} & \textbf{11.19} & \textbf{9.91} & \textbf{11.12} \\
        
        \checkmark & & \checkmark & \underline{15.21} & 28.69 & \underline{14.05} & \textbf{13.12} & \textbf{14.17} & \textbf{11.19} & \underline{10.01} & \underline{11.18} \\
        
        \checkmark & \checkmark & \checkmark & \textbf{15.16} & 27.36 & \underline{14.05} & 13.20 & 14.44 & \underline{11.59} & 10.31 & 11.41 \\
        \bottomrule
    \end{tabular}
    }
    \vspace{-2mm}
\end{table*}

\begin{table}[t]
    \centering
    \caption{Hyperparameters for training the CityWalker model.}
    \label{tab:hyperparameter}
    \resizebox{0.75\linewidth}{!}{
    \begin{tabular}{lc}
    \toprule
    Hyperparameter & Value \\
    \midrule
    \multicolumn{2}{l}{\textbf{CityWalker Model}} \\
    Total \# Parameters & 214M \\
    Trainable \# Parameters & 127M \\
    Image Encoder & DINOv2~\cite{oquab2024dinov} \\
    Backbone Arch. & ViT-B/14 \\
    City Walking Input Res. & $350 \times 630$ \\
    Teleop Input Resolution & $392 \times 392$ \\
    Token Dimension & 768 \\
    Attn. Hidden Dim. & 768 \\
    \# Attention Layers & 16 \\
    \# Attention Heads & 8 \\
    Input Context & 5 \\
    Prediction Horizon & 5 \\
    Input Cord. Repr. & Polar Cord. \\
    Fourier Feat. Freq & 6 \\
    \midrule
    \multicolumn{2}{l}{\textbf{Training}} \\
    \# Epochs & 10 \\
    Batch Size & 32 \\
    Learning Rate & $2\times 10^{-4}$ \\
    Optimizer & AdamW~\cite{loshchilov2018adamw} \\
    LR Schedule & Cosine \\
    Compute Resources & 2 $\times$ H100 \\
    Training Walltime & 30 hours \\
    Fine-tuning LR & $5\times 10^{-5}$ \\
    L1 Loss Weight $\varphi_\text{l1}$ & 1.0 \\
    Ori. Loss Weight $\varphi_\text{ori}$ & 5.0 \\
    Arr. Loss Weight $\varphi_\text{arr}$  &  1.0 \\
    Feat. Loss Weight $\varphi_\text{feat}$ & 0.1 \\
    \bottomrule
    \end{tabular}
    }
    \vspace{-1mm}
\end{table}

\begin{table}[t]
    \centering
    \caption{VLM Results on Offline Data.}
    \label{tab:gpt-result}
    \resizebox{1\linewidth}{!}{
    \begin{tabular}{l|ccc|ccc}
    \toprule
    \multirow{2}{*}{\bf Scenario} & \multicolumn{3}{c|}{\bf GPT-4o~\cite{achiam2023gpt}} & \multicolumn{3}{c}{\bf Ours}  \\
     & $\downarrow$AOE($5$) & $\downarrow$MAOE & $\uparrow$Arrival & $\downarrow$AOE$(5)$ & $\downarrow$MAOE & $\uparrow$Arrival \\
    \midrule
    Mean & 72.22$^\circ$ & 87.39$^\circ$ & 69.38\% & 7.97$^\circ$ & 15.23$^\circ$ & 81.85\% \\
    Turn & 68.61$^\circ$ & 88.02$^\circ$ & 68.66\% & 19.67$^\circ$ & 26.63$^\circ$ & 68.91\% \\
    Cros. & 65.33$^\circ$ & 81.12$^\circ$ & 66.52\% & 5.43$^\circ$ & 14.07$^\circ$ & 75.03\% \\
    Detour & 76.86$^\circ$ & 90.76$^\circ$ & 68.81\% & 8.71$^\circ$ & 13.94$^\circ$ & 78.54\% \\
    Prox. & 75.65$^\circ$ & 95.74$^\circ$ & 66.33\% & 5.54$^\circ$ & 14.32$^\circ$ & 90.64\% \\
    Crowd & 75.85$^\circ$ & 84.88$^\circ$ & 75.47\% & 4.77$^\circ$ & 12.01$^\circ$ & 87.50\% \\
    Other & 71.03$^\circ$ & 83.85$^\circ$ & 70.49\% & 3.67$^\circ$ & 10.40$^\circ$ & 90.19\% \\
    All & 71.51$^\circ$ & 85.03$^\circ$ & 70.04\% & 4.63$^\circ$ & 11.53$^\circ$ & 87.84\% \\
    \bottomrule
    \end{tabular}
    }
\end{table}

\begin{figure}[t]
    \centering
    \includegraphics[width=0.9\linewidth]{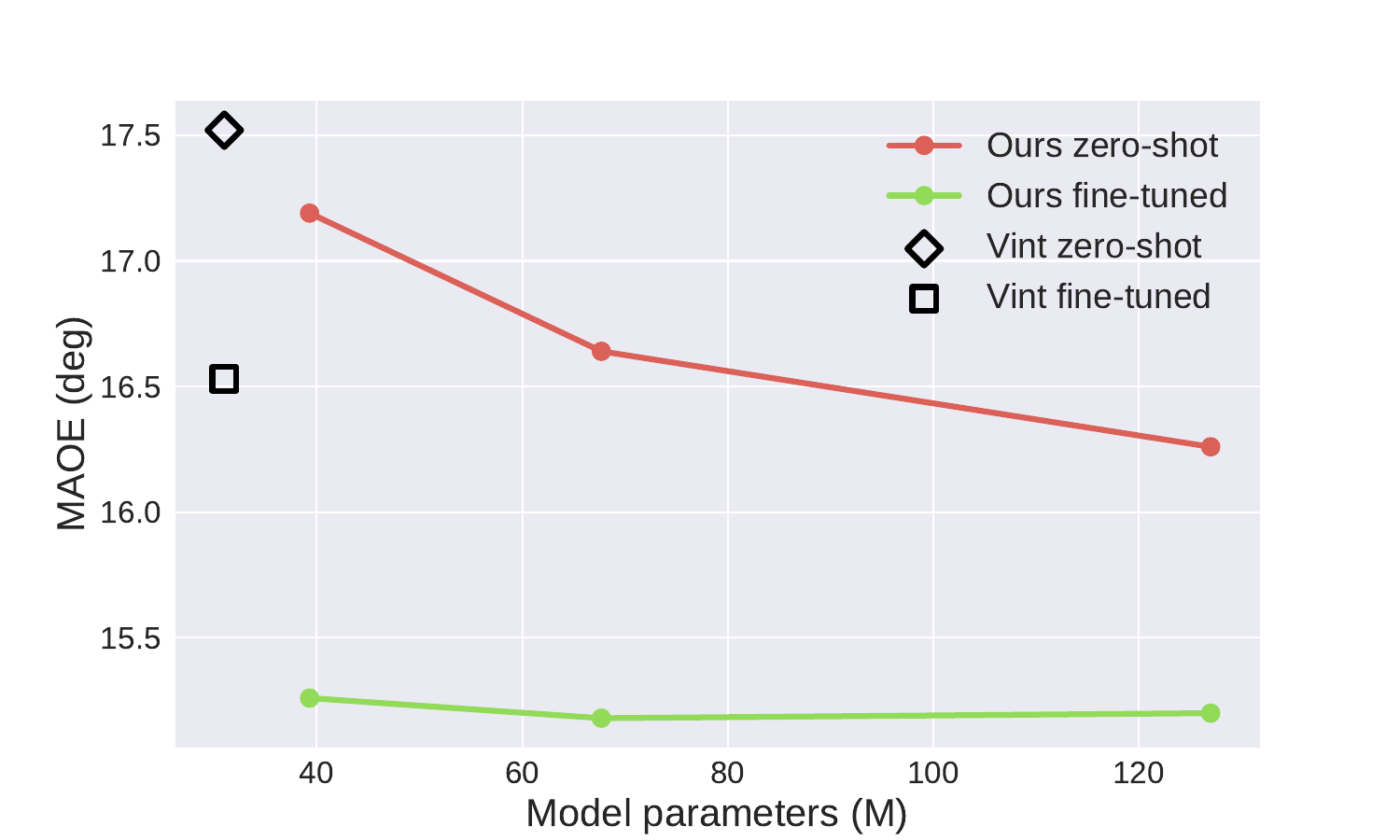}
    \caption{\textbf{Performance and Model Szie}. We show model performance with respect to the size of the model, measured by the number of parameters in the model.}
    \label{fig:params}
    \vspace{-4mm}
\end{figure}

\noindent \textbf{Impact of Model Size}. We also run experiments to discover the impact of model size on navigation performance. This is done by modifying the number of layers in the transformer model. From \cref{fig:params}, we can observe the clear trend that a larger model with more parameters leads to better performance, especially in the zero-shot case. Note that all models in the figure are trained with 2000 hours of video data and we can see a trend of saturation with even larger models. This aligns with the scaling law observed in previous works~\cite{brown2020language,kirillov2023segment,oquab2024dinov,driess2023palm} that a larger model should be accompanied with larger data to produce better results.

\noindent \textbf{Image Backbones}. In \cref{tab:architecture}, we show that our model performance is not sensitive to the choice of image backbones. This makes embodied depolyment very efficient. While our model with DiNOv2 backbone only has 1.7 fps inference speed on a RTX 3060 laptop, this can be boosted to 20 fps by switching to EfficientNetB0 backbone without sacrificing model performance.

\section{More Qualitative Results}

\begin{figure*}[t]
    \centering
    \vspace{-2mm}
    \includegraphics[width=0.85\linewidth]{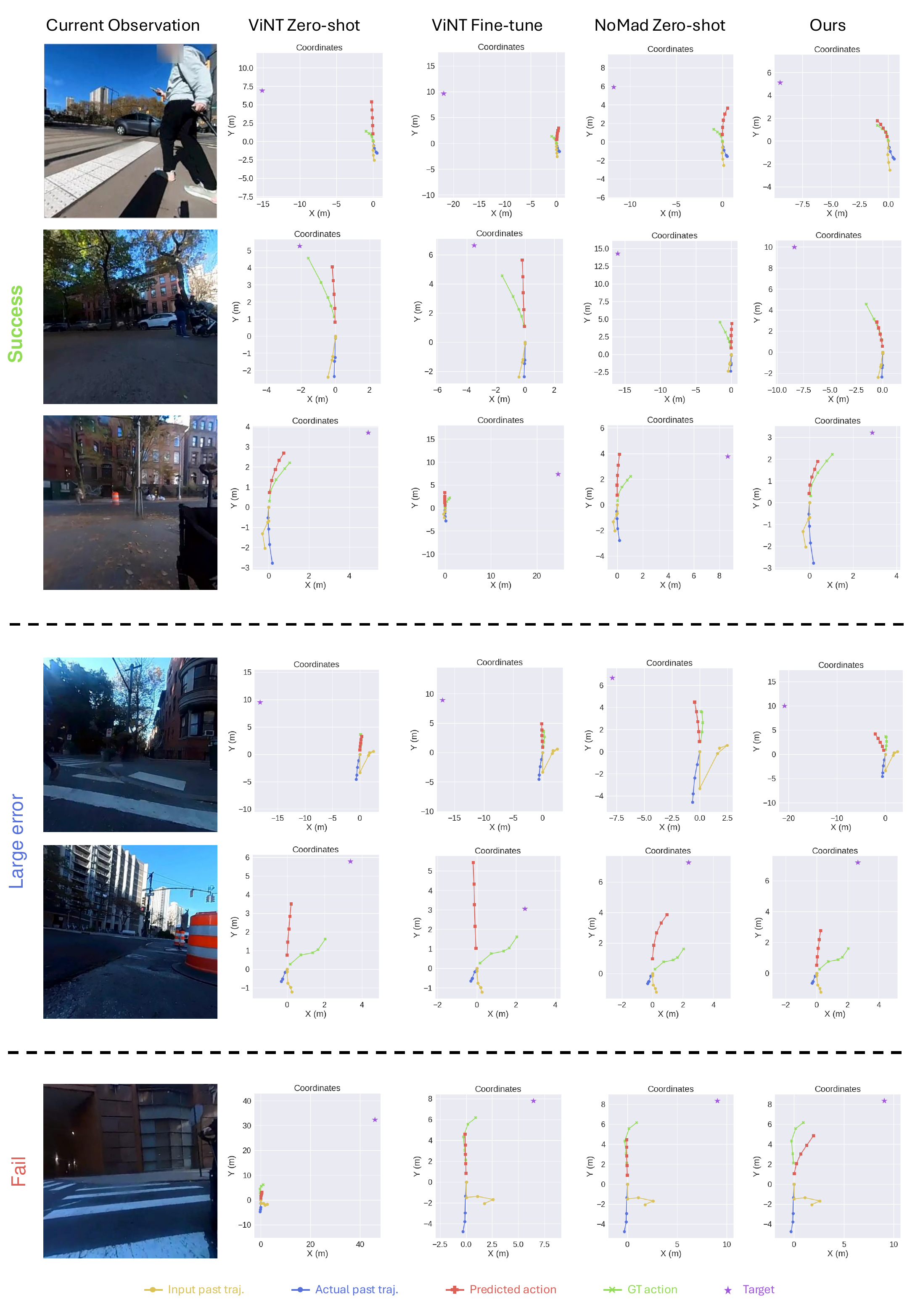}
    \vspace{-4mm}
    \caption{\textbf{More Qualitative Results}. We provide more qualitative results in our offline testing. The results are categorized into success, large error, and fail. Success means the predicted action aligns with ground truth action. Large error mean prediction action does not align with ground truth but still lead to success navigation. Fails cases are those may lead to failed navigation.}
    \label{fig:full-qualitative}
\end{figure*}

In \cref{fig:full-qualitative}, we provide more qualitative resting results on the offline data. We divide the results into three categories. \textcolor{sbgreen}{Success}: predicted action aligns well with ground truth action. \textcolor{sbblue}{Large error}: predicted action does not align with ground truth but may still lead to successful navigation. \textcolor{sbred}{Fail}: predicted action may lead to failed navigation. The most significant observation is that large errors in offline data do not necessarily lead to failure in navigation, due to the multi-modality characteristic of policy learning. For example, in the fifth row, although the ground truth action takes a detour to the right of the traffic drum, the predicted action that goes straight from the left of the drum should also lead to successful navigation.


\begin{table}[t]
    \caption{\textbf{Comparison of backbones and architecture.} All models are \textit{pretrained}
    with 2000 hours of video and fine-tuned with expert data. Both metrics are taking
    the category mean. *Pretrained from ACO~\cite{zhang2022learning}.
    }
    \label{tab:architecture}
    \centering
    \resizebox{\linewidth}{!}{
    \begin{tabular}{l|cccc|c}
        \toprule \textbf{Metric}     & \textbf{EfficientNetB0} & \textbf{ResNet50}           & \textbf{DiNOv2} & \textbf{ResNet34*}        & \textbf{ViNT**}  \\
        \midrule MAOE ($\downarrow$) & 15.33$^{\circ}$         & \underline{15.16$^{\circ}$} & 15.23$^{\circ}$ & \textbf{15.13 $^{\circ}$} & 15.26 $^{\circ}$ \\
        L2 ($\downarrow$)            & 1.11 m                  & 1.15 m                      & 1.12 m          & \underline{1.09 m}        & \textbf{1.08 m}  \\
        \bottomrule
    \end{tabular}
    }
\end{table}

\end{document}